\documentclass[10pt,twocolumn,letterpaper]{article}
\usepackage[pagenumbers]{cvpr}
\usepackage[T1]{fontenc}
\usepackage[utf8]{inputenc}
\usepackage{textcomp,amsmath,amssymb,booktabs,array,tabularx,makecell,multirow}
\usepackage{tikz}
\usetikzlibrary{arrows.meta,positioning,fit,backgrounds,calc}
\usepackage{enumitem}
\usepackage{caption}
\usepackage{microtype}
\usepackage{placeins}
\definecolor{cvprblue}{rgb}{0.21,0.49,0.74}
\usepackage[breaklinks,colorlinks,allcolors=cvprblue]{hyperref}

\def\confName{CVPR}
\def\confYear{2027}

\title{Measuring the Planner in Latent World Model Evaluation}
\author{Kaixuan Yin$^{1*}$\quad Yanshen Xie$^{1*}$\quad Shicheng Yin$^{1}$\quad Peng Zou$^{1}$\quad Yang Liu$^{1,2\dagger}$\\
$^{1}$Sun Yat-sen University\quad $^{2}$X-Era AI Lab\\
{\small \{yinkx3, xieysh33, yinshch5, zoup7\}@mail2.sysu.edu.cn,\ \ liuy856@mail.sysu.edu.cn}\\
{\small Code: \href{https://github.com/HCPLab-SYSU/RM-CEM}{\textcolor{black}{github.com/HCPLab-SYSU/RM-CEM}}}}
\makeatletter\@namedef{ver@eso-pic.sty}{2020/10/14}\makeatother
\usepackage{pdfpages}
\begin{document}
\twocolumn[{\renewcommand\twocolumn[1][]{#1}\maketitle\begin{center}\vspace{-1.2em}\includegraphics[width=\textwidth]{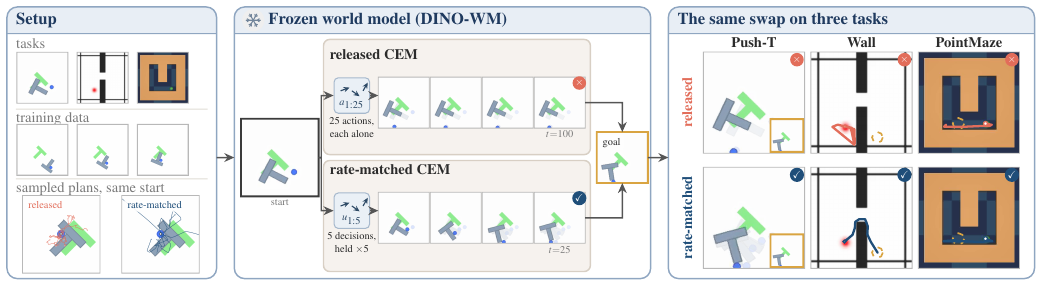}\captionof{figure}{A failed episode may be the planner's rather than the model's. \emph{Center:} with the same frozen world model (DINO-WM), cost, and rollout budget, the released CEM samples every primitive action independently and fails, while rate-matched CEM samples one decision per model step, holds it for the actions that step reads, and reaches the goal. \emph{Left:} the tasks, training data, and plans sampled by the two proposals from one start state. \emph{Right:} the same swap on Push-T, Wall, and PointMaze, with executed paths on the start frame. $\times$ and \checkmark{} mark failure and success; the corner inset or dashed circle marks the goal (the green T in Push-T is part of the environment, not the goal).}\label{fig:teaser}\end{center}\vspace{0.6em}}]
{\let\thefootnote\relax\footnotetext{$^{*}$Equal contribution.\quad $^{\dagger}$Corresponding author.}}
\begin{abstract}Latent world models such as DINO-WM, DDP-WM, and JEPA-WMs are ranked by how often planning with them succeeds on Push-T, Wall, and PointMaze, yet a failed episode may be the planner's rather than the model's. We bound the planner's share: a planner that uses only the model, its cost, and the evaluation budget can only avoid failures that the model does not force. Such a planner follows from how these models are trained. They read actions in blocks of five, and their training actions vary mostly from block to block, while the released cross-entropy method (CEM) samples every action independently, giving the block means one fifth of its exploration. Rate-matched CEM searches one decision per block, with no parameter. It succeeds more often in all nine model--benchmark settings, and its results exceed the best official ones on every benchmark. The bound places 55\% of the reported failures with the search (38\% at 95\% confidence). Interpolated plans and late refinement keep the gain, and colored noise does not reproduce it on Push-T and Wall. Over 22 settings that include four further world models and gradient-based planning, held decisions never lower success significantly; the one significant loss, under a planner that executes single samples, is traced to plans that leave the training range.\end{abstract}

\section{Introduction}

Latent world models plan in the feature space of a pretrained encoder. DINO-WM~\cite{ref44}, DDP-WM~\cite{ref42}, and JEPA-WMs~\cite{ref38} predict in the features of a frozen DINOv2 encoder, plan with the cross-entropy method (CEM)~\cite{ref34} inside model predictive control, and are compared by their planning success on Push-T, Wall, and PointMaze. These comparisons guide design. DDP-WM, the current state of the art, replaces the predictor of DINO-WM while keeping its environments, data, and core CEM parameters, and JEPA-WMs ranks design choices for JEPA world models by their success under CEM.

A success rate measures a model and a planner together (Figure 1). Following machine translation~\cite{ref13,ref37}, we call a failure a search error when another planner, given the same model, cost, and budget, would avoid it, and a model error otherwise. The distinction matters most near the ceiling, where DINO-WM and DDP-WM now operate. Their remaining failures are few, and a comparison between them compares models only to the extent that those failures are model errors. For latent world models, the split has not been measured.

Any planner that uses nothing but the model, its cost, and the budget gives a bound on it. Whatever such a planner avoids was not forced by the model, so its failure rate bounds model error from above, and its improvement over the released planner bounds search error from below (Section 2). The bound costs one further evaluation and holds whatever the reason for the improvement. What it needs is a planner that finds plans the released one misses.

The training of these models points to such a planner. DINO-WM and DDP-WM are trained with a frameskip, so one predictor step reads a block of five primitive actions. In the Push-T training data, neighboring actions correlate at 0.87, and 79\% of the action variance lies in the block means. The released planner nonetheless samples the five actions of a block independently. We show that this gives the block means one fifth of the exploration at every iteration, however the proposal is refitted or rescaled (Proposition 1). Rate-matched CEM samples one decision per block instead. Its block length comes from training, so it has no parameter, and it keeps the released model, cost, budget, and controller.

Our findings are as follows.

\begin{itemize}[leftmargin=*,nosep]\item On the standard suite, rate-matched CEM succeeds more often than the released planner in all nine model--benchmark settings, and its results exceed the best official results on every benchmark. The bound places 55\% of the reported failures with the search, at least 38\% at 95\% confidence, and 74\% on the navigation tasks. On Wall, the change of planner removes 11 of DINO-WM's 13 failures, and the change from DINO-WM's predictor to DDP-WM's removes 2 (Section 4).\end{itemize}

\begin{itemize}[leftmargin=*,nosep]\item The gain needs neither repeated actions nor a small final search space. Colored noise in the style of iCEM~\cite{ref33}, which raises low-frequency exploration without deciding per block, does not reproduce it on Push-T and Wall. Searching decisions at the step-mean scale of the released proposal removes it (Section 5).\end{itemize}

\begin{itemize}[leftmargin=*,nosep]\item The direction carries over beyond the standard suite. On four further world models, under CEM, EB-CEM, and gradient descent, held decisions raise success in ten of thirteen settings, significantly on UMaze, and lower it significantly in none. Under NG, a CMA-ES planner that executes a single sample, they help on Wall and PointMaze and hurt on Push-T, where 35.4\% of the held five-step displacements leave the training range, against 0.7\% under CEM. This measures directly the second way in which a planner fails, and it shows which planners held decisions suit (Section 6).\end{itemize}

We contribute (i) a split of reported failure into model error and search error, bounded by any planner in a precisely defined class; (ii) rate-matched CEM, a parameter-free planner derived from how per-action sampling explores block-input models, with reference results for three published models; and (iii) a controlled evaluation across seven world models, four planners, and four environments, with failure counts and exact tests.

\section{Search error and model error}

Let M be a world model together with its planning cost, and let B be the evaluation budget, the rollouts per planning round and the replanning schedule. The planner class \ensuremath{\Pi}(M, B) contains every planner whose plan depends only on the costs, or cost gradients, that M assigns to the candidates it evaluates within B; the observation o and the goal \ensuremath{o_g} enter only through their encodings E(o) and E(\ensuremath{o_g}). A member is one algorithm applied unchanged to every task, and a randomized member is scored by its expected failure rate. Policies, simulators, hand-designed controllers on raw observations, and second learned models are excluded, because they bring knowledge that M lacks (Appendix F).

With F(M, P) the closed-loop failure rate of planner P, the model error of M is the lowest failure rate in the class,
\begin{equation*}E_{\mathrm{model}}(M)=\inf_P F(M,P)\ \text{over}\ \Pi(M,B),\end{equation*}
and the search error of the released planner \ensuremath{P_0} is the remainder:

\begin{equation*}F(M,P_0)=E_{\mathrm{model}}(M)+E_{\mathrm{search}}(M,P_0).\tag{1}\end{equation*}

The infimum is not observable, but every member \ensuremath{P_1} gives

\begin{equation*}\begin{aligned}E_{\mathrm{model}}(M)&\le F(M,P_1)\quad\text{and}\\ E_{\mathrm{search}}(M,P_0)&\ge F(M,P_0)-F(M,P_1).\end{aligned}\tag{2}\end{equation*}

Three remarks fix what (2) says. First, model error belongs to M together with its cost and budget, not to the predictor alone: it is the failure that no member can avoid under the evaluation as it stands. Second, the bound holds whatever the reason \ensuremath{P_1} does better and needs no per-episode diagnosis; a stronger member only raises it, so the bound is as useful as the plans \ensuremath{P_1} finds that \ensuremath{P_0} misses. Third, a planner can add failure in two ways. It can miss a plan that M ranks well and that works, the search error of machine translation, or it can query M where M was not trained and return a plan that M overrates~\cite{ref23,ref22}. Another member can avoid both, so both count as search error and (2) bounds their sum. Section 6.1 measures the second way directly.

\section{Block-input models and per-action sampling}

\subsection{Models and the released planner}

All three models encode an observation with a frozen DINOv2 network E~\cite{ref32} and advance the encoding with a learned predictor f. DINO-WM and DDP-WM train f with a frameskip: one step takes the latent state and a block \ensuremath{a_h} = (\ensuremath{a_{h,1}}, ..., \ensuremath{a_{h,k}}) of k = 5 primitive actions in \ensuremath{\mathbb R^d} and returns the latent state k environment steps later. A plan over H model steps is an array a \ensuremath{\in} \ensuremath{\mathbb R^{H\times k\times d}}, with 50 entries on Push-T, where d = 2 and H = 5. The official cost C(a) compares the predicted latent state with E(\ensuremath{o_g}).

The released planner runs CEM inside model predictive control. Each of T iterations samples N plans, drawing every primitive action independently from a proposal that starts at standard deviation one in standardized units; it rolls the plans out through f, ranks them by C, and refits the proposal to the best K. On Push-T, DINO-WM uses N = 300, K = 30, and T = 30, or 9,000 rollouts per planning round, and the controller executes the plan and replans, for at most three planning rounds per task.

\subsection{The training actions}

Write
\begin{equation*}\bar a_h=\frac1k\sum_j a_{h,j}\quad\text{and}\quad\delta_{h,j}=a_{h,j}-\bar a_h\end{equation*}
for the step mean of a block and for its within-step deviations. Since the deviations sum to zero,

\begin{equation*}\sum_j\|a_{h,j}\|^2=k\|\bar a_h\|^2+\sum_j\|\delta_{h,j}\|^2.\tag{3}\end{equation*}

The step subspace S, the plans without within-step deviation, is the range of the lift
\begin{equation*}L:\mathbb R^{H\times d}\to\mathbb R^{H\times k\times d}\quad\text{with}\quad L(u)_{h,j}=u_h.\end{equation*} The Push-T training set of DINO-WM and DDP-WM, 18685 trajectories or about 460k five-action chunks (Appendix C.3), keeps 79\% of its variance in the step means in standardized units, with a step-mean standard deviation of 0.89; replacing every chunk by its mean discards 21\%. A stationary first-order process with the measured neighbor correlation of 0.87 would keep 81\% (Appendix A.4), and a random subspace of the dimension of S keeps 20\% on average (Appendix A.5). The training actions lie close to S. Plans in S also solve the tasks. On the 21 validation trajectories of the same Push-T data, we replaced each block of recorded actions by its mean and executed the result. It reaches the state recorded 25 steps later in all 21 trajectories under the official success criterion, and the same holds 50 and 75 steps later. The model scores these block-averaged plans as it scores the recorded ones, with a median cost of 0.13 for both at 25 steps against 1.93 for random action sequences. The bounds of Section 4 do not use these statistics.

\begin{figure*}[t]\centering

\includegraphics[width=\textwidth]{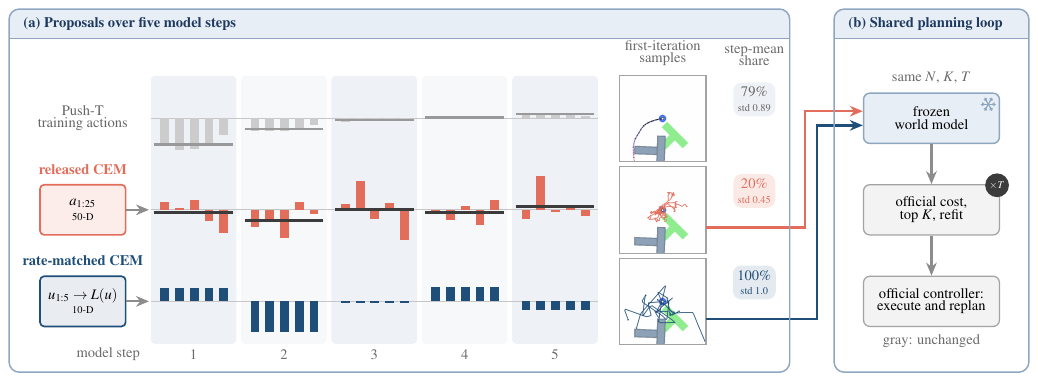}

\setcounter{figure}{1}

\caption{(a) One action coordinate over the 25 primitive actions of five model steps, in standardized units: Push-T training actions, a draw of the released CEM proposal, and a draw of rate-matched CEM; horizontal lines mark step means. Right of each row: a Push-T path from the training data or from first-iteration samples, and the share of exploration on the step means with the standard deviation of a step mean (Table 1). (b) The planning loop that both planners share; gray boxes are unchanged.}\end{figure*}

\subsection{How per-action sampling explores}

Applying (3) to the sampling noise \ensuremath{\varepsilon_{h,j}} = \ensuremath{a_{h,j}} \ensuremath{-} \ensuremath{\mu_{h,j}} around the proposal means gives the following.

Proposition 1. Let the noise terms \ensuremath{\varepsilon_{h,1}}, ..., \ensuremath{\varepsilon_{h,k}} of a block be independent with zero mean and covariances \ensuremath{\Sigma_{h,1}}, ..., \ensuremath{\Sigma_{h,k}}. Then
\begin{equation*}\begin{aligned}\mathrm{E}\big[k\|\bar\varepsilon_h\|^2\big]&=\frac1k\sum_j\operatorname{tr}\Sigma_{h,j}\quad\text{and}\\ \mathrm{E}\Big[\sum_j\|\varepsilon_{h,j}-\bar\varepsilon_h\|^2\Big]&=\frac{k-1}{k}\sum_j\operatorname{tr}\Sigma_{h,j}.\end{aligned}\end{equation*}

The share of exploration on the step means is 1/k whatever the means and variances (Appendix A.1). A refit changes both but keeps the actions of a block independent, so at k = 5 the released planner gives the step means one fifth of its exploration at every iteration, where the training actions keep 79\% of their variance. From the first iteration, a step-mean coordinate reaches the spread of the data, |\ensuremath{\bar a_h}| \ensuremath{\ge} 0.89, with probability of about 4.7\%, against 37\% for one decision per step drawn at the same scale (Table 1, Appendix A.3).

Neither obvious adjustment changes this. Scaling the proposal scales both parts: giving the step means standard deviation 1.0 would take \ensuremath{\sqrt5} per primitive action and leave within-step deviations of standard deviation 2. Raising the share requires correlation between the actions of a block (Appendix A.4), which colored noise provides, and Section 5.2 tests it. For idealized costs, Appendix A adds that one refit moves a per-action plan 1/\ensuremath{\sqrt k} as far along a cost linear in the step means as a per-step search does (Proposition 3), and that selection on a cost of the step means alone leaves within-step deviations unshaped (Proposition 2).

\begin{table}[t]
\centering\footnotesize\setlength{\tabcolsep}{3pt}
\setcounter{table}{0}
\caption{Exploration per action coordinate at k = 5, in standardized units, at the first CEM iteration; the official share holds at every iteration.}
\begin{tabular}{@{}>{\raggedright\arraybackslash}p{0.3000\dimexpr\linewidth-6\tabcolsep\relax}>{\raggedright\arraybackslash}p{0.2333\dimexpr\linewidth-6\tabcolsep\relax}>{\raggedright\arraybackslash}p{0.2333\dimexpr\linewidth-6\tabcolsep\relax}>{\raggedright\arraybackslash}p{0.2333\dimexpr\linewidth-6\tabcolsep\relax}@{}}
\toprule
 & std of step mean & share on step means & P(|\ensuremath{\bar a_h}| \ensuremath{\ge} 0.89) \\
\midrule
official CEM & 0.45 & 20\% & 4.7\% \\
Push-T training actions & 0.89 & 79\% & -- \\
rate-matched CEM & 1.0 & 100\% & 37\% \\
\bottomrule\end{tabular}
\end{table}

\subsection{Rate-matched CEM}

Rate-matched CEM keeps the official loop and samples in S (Algorithm 1). It draws decisions u \ensuremath{\in} \ensuremath{\mathbb R^{H\times d}}, one per model step, lifts each to L(u), rolls it out through the frozen predictor, ranks it by the official cost, and refits its proposal over u to the K elites. With the official N, K, and T it spends the same 9,000 rollouts per round, and the official controller executes and replans as before.

\begin{quote}\small\hrule\vspace{4pt}Algorithm 1: Rate-matched CEM, one planning round.\par Input: observation o, goal \ensuremath{o_g}, frozen E and f, official cost C, frameskip k, horizon H, official budget (N, K, T).\par 1: initialize the proposal over u \ensuremath{\in} \ensuremath{\mathbb R^{H\times d}} as the official planner initializes its own, with standard deviation 1\par 2: repeat T times:\par 3:   sample \ensuremath{u^{(1)},\dots,u^{(N)}} from the proposal\par 4:   lift each sample, \ensuremath{a^{(n)}_{h,j}=u^{(n)}_h} for j = 1, ..., k\par 5:   roll every \ensuremath{a^{(n)}} out through f from E(o) and evaluate C\par 6:   refit the proposal on the u of the K lowest-cost samples\par 7: pass the lifted plan to the official controller\vspace{4pt}\hrule\end{quote}

The planner has no parameter. S is fixed by the training frameskip and everything else is the released planner's, so rate-matched CEM is a member of \ensuremath{\Pi}(M, B) chosen without looking at the evaluation. On Push-T it searches 10 of the official 50 dimensions, and against Table 1 three things change at once: the step means receive all of the exploration instead of a fifth, they start at standard deviation 1.0 instead of 0.45, and the search has a fifth of the dimensions. The bound (2) does not depend on which of these matters; Section 5 tests what the gain does not require, and Section 5.3 isolates the scale.

\section{The standard suite}

\subsection{Setup}

DINO-WM and DDP-WM run with their official implementations and evaluation pipelines. Their predictors are far apart: a dense transformer over all image patches in DINO-WM and, in DDP-WM, sparse primary dynamics with a low-rank correction of the background. JEPA-WMs runs from its official public weights and configuration, with CEM under the L2 cost, and rate-matched CEM holds each decision for that model's frameskip. In every comparison, \ensuremath{P_0} is the released CEM planner of the same model and \ensuremath{P_1} is rate-matched CEM, run on the same tasks. Push-T~\cite{ref11} asks for a T-shaped block to be pushed into a goal pose; Wall~\cite{ref44} and PointMaze~\cite{ref12} are navigation tasks. These three are how JEPA world models are currently evaluated for planning, and few works report others. DINO-WM and DDP-WM use three seeds (99, 1, and 2) of 50 tasks, 150 episodes per setting, and JEPA-WMs uses one seed of 96 episodes; the official evaluation decides success. We report failed episodes beside success rates and two-sided exact Cochran--Mantel--Haenszel tests stratified by benchmark~\cite{ref9,ref27} (Appendix B).

\subsection{All nine settings improve}

Rate-matched CEM succeeds more often than the released planner in all nine settings (Table 2), although it searches a subset of the official plans with the same rollouts. Within each model, the exact test over its three benchmarks rejects equal failure probabilities, with p = 0.004 for DINO-WM, p = 0.015 for DDP-WM, and p < \ensuremath{10^{-4}} for JEPA-WMs. Single cells, such as DINO-WM on Push-T with 12 failures against 8, are not significant alone; the evidence is their agreement. The rate-matched rows serve as reference results for the released checkpoints. On Push-T, Wall, and PointMaze their best values, 94.7\%, 98.7\%, and 100\%, exceed the best official results of DINO-WM and DDP-WM, which are 92.0\%, 92.7\%, and 98.0\%. DDP-WM, the current state of the art, improves on all three benchmarks, and JEPA-WMs, which the official planner leaves furthest from the ceiling, gains 13.5, 15.6, and 10.4 points.

\begin{table*}[t]
\centering\footnotesize\setlength{\tabcolsep}{3pt}
\setcounter{table}{1}
\caption{Success rate (\%) with failed episodes in parentheses; 150 episodes per setting for DINO-WM and DDP-WM, 96 for JEPA-WMs.}
\begin{tabular}{@{}>{\raggedright\arraybackslash}p{0.3000\dimexpr\linewidth-10\tabcolsep\relax}>{\raggedright\arraybackslash}p{0.1400\dimexpr\linewidth-10\tabcolsep\relax}>{\raggedright\arraybackslash}p{0.1400\dimexpr\linewidth-10\tabcolsep\relax}>{\raggedright\arraybackslash}p{0.1400\dimexpr\linewidth-10\tabcolsep\relax}>{\raggedright\arraybackslash}p{0.1400\dimexpr\linewidth-10\tabcolsep\relax}>{\raggedright\arraybackslash}p{0.1400\dimexpr\linewidth-10\tabcolsep\relax}@{}}
\toprule
Model & Planner & Push-T & Wall & PointMaze & Failures \\
\midrule
DINO-WM & official & 92.0 (12) & 91.3 (13) & 98.0 (3) & 28 \\
DINO-WM & rate-matched & 94.7 (8) & 98.7 (2) & 100.0 (0) & 10 \\
DDP-WM & official & 87.3 (19) & 92.7 (11) & 96.0 (6) & 36 \\
DDP-WM & rate-matched & 90.7 (14) & 98.7 (2) & 98.7 (2) & 18 \\
JEPA-WMs & official & 64.6 (34) & 75.0 (24) & 86.5 (13) & 71 \\
JEPA-WMs & rate-matched & 78.1 (21) & 90.6 (9) & 96.9 (3) & 33 \\
\bottomrule\end{tabular}
\end{table*}

\begin{table*}[t]
\centering\footnotesize\setlength{\tabcolsep}{3pt}
\setcounter{table}{2}
\caption{Failures under both planners and the bounds (2): search error as a share of reported failures, model error as a failure rate.}
\begin{tabular}{@{}>{\raggedright\arraybackslash}p{0.2000\dimexpr\linewidth-10\tabcolsep\relax}>{\raggedright\arraybackslash}p{0.1600\dimexpr\linewidth-10\tabcolsep\relax}>{\raggedright\arraybackslash}p{0.1600\dimexpr\linewidth-10\tabcolsep\relax}>{\raggedright\arraybackslash}p{0.1600\dimexpr\linewidth-10\tabcolsep\relax}>{\raggedright\arraybackslash}p{0.1600\dimexpr\linewidth-10\tabcolsep\relax}>{\raggedright\arraybackslash}p{0.1600\dimexpr\linewidth-10\tabcolsep\relax}@{}}
\toprule
Group & Episodes & Official & Rate-matched & Search error, at least & Model error, at most \\
\midrule
DINO-WM & 450 & 28 & 10 & 64\% & 2.2\% \\
DDP-WM & 450 & 36 & 18 & 50\% & 4.0\% \\
JEPA-WMs & 288 & 71 & 33 & 54\% & 11.5\% \\
Push-T & 396 & 65 & 43 & 34\% & 10.9\% \\
Wall and PointMaze & 792 & 70 & 18 & 74\% & 2.3\% \\
All nine settings & 1,188 & 135 & 61 & 55\% & 5.1\% \\
\bottomrule\end{tabular}
\end{table*}

\subsection{The bound}

Table 3 and Figure 3 apply (2) with rate-matched CEM alone as \ensuremath{P_1}. Over 1,188 episodes per planner, failures fall from 135 to 61, so the lower bound on search error is 55\% of the reported failures, with a one-sided 95\% limit of 38\%; DINO-WM and DDP-WM alone, over three seeds each, give 56\%. Model error is at most 2.2\%, 4.0\%, and 11.5\% of episodes for the three models. The benchmark matters more than the model. On Wall and PointMaze, failures fall from 70 to 18, and the bound covers 74\% of them, at least 53\% at 95\% confidence. Push-T has the smallest share in every model (Appendix D) and holds 43 of the 61 failures that remain; how these divide is left to stronger members of the class, which can only raise the bound.

\emph{On Push-T, the failures that rate-matched CEM avoids end with the pusher out of place.} Push-T counts an episode as solved only when the pusher and the T-block are both close to their goal positions and the block is within 20 degrees of its goal angle. We estimated the final poses in the JEPA-WMs episodes from the evaluation videos, and the estimates agree with the official outcome in 95 of 96 episodes under each planner. In 18 of the 34 failures of the released planner, the T-block ends within 20 units and 20 degrees of its goal and the pusher does not. Rate-matched CEM leaves 5 failures of this kind among its 21. The other failures number 16 under both planners, so the net change of 13 failures lies in this kind alone.

\begin{figure}[t]\centering

\includegraphics[width=\linewidth]{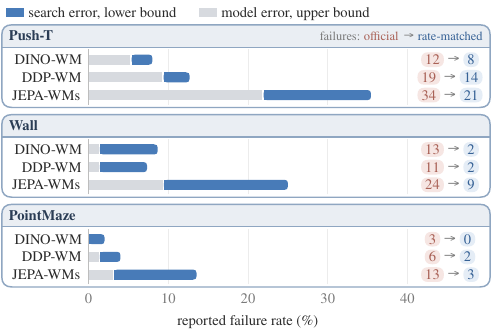}

\setcounter{figure}{2}

\caption{For each model and benchmark, the failure rate of the official planner, split by (2) into the lower bound on search error (blue) and the upper bound on model error (gray).}\end{figure}

\subsection{Replanning rounds}

The budget B includes the replanning schedule. The official controller plans, executes the plan, and plans again from the new observation, for up to three rounds, so a plan that the search got wrong can be repaired in a later round. Figure 4 counts the failures left after each round on Wall and PointMaze, for DINO-WM and DDP-WM, 600 episodes per planner. \emph{Rate-matched CEM needs fewer rounds.} After the first round the released planner fails in 179 episodes and rate-matched CEM in 110. After two rounds rate-matched CEM fails in 19, fewer than the 33 failures that the released planner leaves after all three, and this holds in each of the four settings. The official rounds thus repair part of what the released search misses, and the bound of Section 4.3, read after the third round, does not count what they repair. \emph{On Push-T, replanning changes little.} Over the three rounds the Push-T failures of DDP-WM fall only from 23 to 19 of 150 under the released planner and from 20 to 14 under rate-matched CEM.

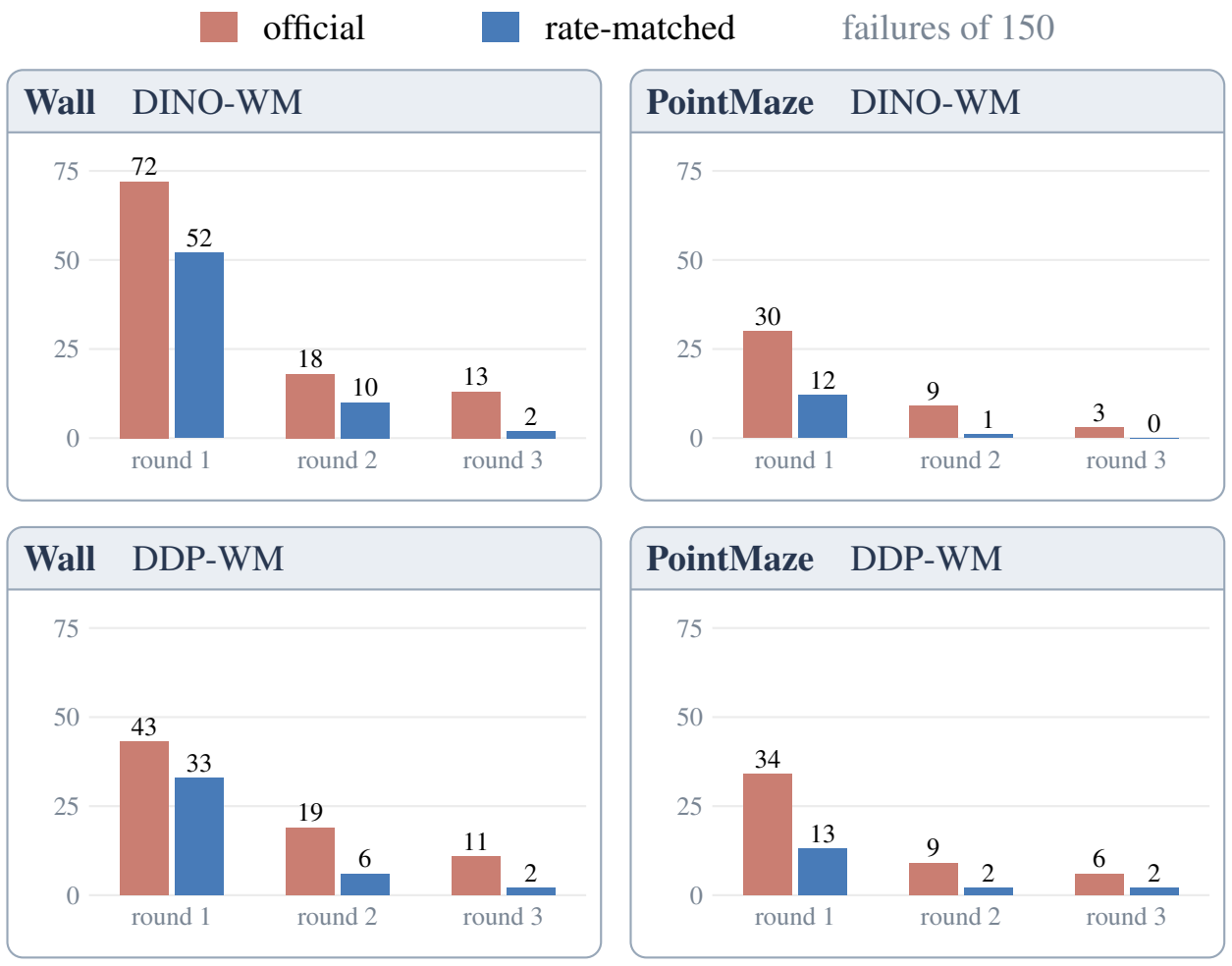
\begin{figure}[t]\centering
\definecolor{rpoff}{rgb}{0.793,0.494,0.446}\definecolor{rprm}{rgb}{0.283,0.482,0.721}\definecolor{rphdr}{rgb}{0.919,0.934,0.952}\definecolor{rpbrd}{rgb}{0.595,0.654,0.723}\definecolor{rpgrid}{rgb}{0.925,0.926,0.925}\definecolor{rpax}{rgb}{0.474,0.523,0.577}\definecolor{rpnavy}{rgb}{0.152,0.210,0.308}
\newcommand{\rpanel}[9]{\begin{scope}[shift={(#1,#2)}]
\begin{scope}\clip[rounded corners=3pt] (0,0) rectangle (4.0,2.9);\fill[rphdr] (0,2.48) rectangle (4.0,2.9);\end{scope}
\draw[rpbrd,rounded corners=3pt] (0,0) rectangle (4.0,2.9);\draw[rpbrd] (0,2.48)--(4.0,2.48);
\node[anchor=west,text=rpnavy,inner sep=0pt] at (0.1,2.69) {#3};
\foreach \v in {0,25,50,75}{\draw[rpgrid] (0.55,{0.42+0.024*\v})--(3.9,{0.42+0.024*\v});\node[anchor=east,text=rpax,font=\tiny,inner sep=1pt] at (0.53,{0.42+0.024*\v}) {\v};}
\foreach \k/\fo/\fr in {1/#4/#7,2/#5/#8,3/#6/#9}{\pgfmathsetmacro{\cx}{0.55+(\k-0.5)*1.117}
\fill[rpoff] ({\cx-0.35},0.42) rectangle ({\cx-0.02},{0.42+0.024*\fo});\fill[rprm] ({\cx+0.02},0.42) rectangle ({\cx+0.35},{0.42+0.024*\fr});
\node[font=\tiny,anchor=south,inner sep=1pt] at ({\cx-0.185},{0.42+0.024*\fo}) {\fo};\node[font=\tiny,anchor=south,inner sep=1pt] at ({\cx+0.185},{0.42+0.024*\fr}) {\fr};
\node[font=\tiny,text=rpax,anchor=north,inner sep=1pt] at (\cx,0.38) {round \k};}
\end{scope}}
\begin{tikzpicture}[x=1cm,y=1cm,font=\scriptsize]
\fill[rpoff] (1.3,6.17) rectangle (1.55,6.37);\node[anchor=west] at (1.6,6.27) {official};
\fill[rprm] (3.2,6.17) rectangle (3.45,6.37);\node[anchor=west] at (3.5,6.27) {rate-matched};
\node[anchor=west,text=rpax] at (5.5,6.27) {failures of 150};
\rpanel{0}{3.08}{\textbf{Wall}\quad DINO-WM}{72}{18}{13}{52}{10}{2}
\rpanel{4.2}{3.08}{\textbf{PointMaze}\quad DINO-WM}{30}{9}{3}{12}{1}{0}
\rpanel{0}{0}{\textbf{Wall}\quad DDP-WM}{43}{19}{11}{33}{6}{2}
\rpanel{4.2}{0}{\textbf{PointMaze}\quad DDP-WM}{34}{9}{6}{13}{2}{2}
\end{tikzpicture}
\setcounter{figure}{3}
\caption{Failures left after each planning round out of 150 episodes, under the released planner and under rate-matched CEM, with the official controller and its three rounds. The third round gives the failure counts of Table 2.}\label{fig:rounds}\end{figure}

\begin{table*}[t]
\centering\footnotesize\setlength{\tabcolsep}{3pt}
\setcounter{table}{3}
\caption{Searches over decisions on DDP-WM, success rate (\%) with failed episodes in parentheses, 150 episodes per setting; ``--'' was not run.}
\begin{tabular}{@{}>{\raggedright\arraybackslash}p{0.3000\dimexpr\linewidth-10\tabcolsep\relax}>{\raggedright\arraybackslash}p{0.1400\dimexpr\linewidth-10\tabcolsep\relax}>{\raggedright\arraybackslash}p{0.1400\dimexpr\linewidth-10\tabcolsep\relax}>{\raggedright\arraybackslash}p{0.1400\dimexpr\linewidth-10\tabcolsep\relax}>{\raggedright\arraybackslash}p{0.1400\dimexpr\linewidth-10\tabcolsep\relax}>{\raggedright\arraybackslash}p{0.1400\dimexpr\linewidth-10\tabcolsep\relax}@{}}
\toprule
Environment & official & held & interpolated & held + late & interpolated + late \\
\midrule
Push-T & 87.3 (19) & 90.7 (14) & 91.3 (13) & 91.3 (13) & 92.7 (11) \\
Wall & 92.7 (11) & 98.7 (2) & 98.7 (2) & 98.7 (2) & 98.7 (2) \\
PointMaze & 96.0 (6) & 98.7 (2) & 98.7 (2) & -- & 98.7 (2) \\
\bottomrule\end{tabular}
\end{table*}

\subsection{Planner and predictor}

DINO-WM and DDP-WM share the encoder, environments, data, and core CEM parameters, so their official rows differ in the predictor alone, and the same tasks set a change of predictor beside a change of search. The change of search removes 18 failures from each model, 4.0 points of failure rate, significantly within each. The predictors differ by 8 failures under either planner, 28 against 36 and 10 against 18, which is not significant, and their order changes between benchmarks. On Wall, the predictor of DDP-WM removes 2 of DINO-WM's 13 failures, and rate-matched CEM with the dense predictor removes 11. On this suite the planner is a factor of the same order as the predictor. We therefore suggest reporting failed episodes beside success rates, and running rate-matched CEM beside the released planner, which turns every reported setting into a bound (2) for one further evaluation.

\section{What the gain does not require}

Rate-matched CEM repeats each action five times, ends its search in a small space, and concentrates exploration at low frequency. Each could explain the gain in familiar terms. The bound does not depend on the explanation, but the explanation decides what practitioners should take from the result.

\subsection{Repeated actions and a small final space}

Two variants on DDP-WM remove the first two features (Table 4). Interpolation forms the primitive actions by linear interpolation between neighboring decisions, as spline-based predictive sampling does~\cite{ref21}, so that no action is repeated. Late refinement searches decisions for the first 15 iterations and every primitive action afterwards, ending in the full official space. On Push-T this splits the 30 official iterations in half; on Wall and PointMaze the official search runs 10 iterations, so the switch is never reached and late refinement coincides with the variant without it (supplementary material, Appendix C.2). All four combinations spend the official budget.

All four beat the official planner wherever they ran, by 3.4 to 5.4 points on Push-T, 6.0 on Wall, and 2.7 on PointMaze, and stay within two points of one another. Interpolated and held plans differ by 0.6 and 1.4 points on Push-T under the two schedules and not at all on Wall and PointMaze, so repeating actions is not what helps. On Push-T, late refinement reaches 91.3\% in the same 50-dimensional space, with the same budget, in which the official planner reaches 87.3\%. The two differ only in the order in which the budget is spent, so ending in a small space is not what helps either. Freeing the within-step directions late brings no failures back on Push-T, where they fall from 14 to 13 and from 13 to 11.

\subsection{Correlated noise}

iCEM samples colored noise whose power falls with frequency as \ensuremath{1/f^\beta}~\cite{ref33}, and path-integral planners filter their noise or sample it at low frequency~\cite{ref31,ref39}. Such correlation lifts the step-mean share above 1/k (Appendix A.4) while every primitive action stays its own coordinate, so it shows how far a higher share goes by itself. We give the official CEM of JEPA-WMs the colored noise of iCEM alone, at \ensuremath{\beta} = 2.5, on the tasks of Table 2 (Figure 5). Out of 96 episodes it solves 53 on Push-T, 66 on Wall, and 94 on PointMaze, against 62, 72, and 83 for the official proposal and 75, 87, and 93 for rate-matched CEM. Paired task by task with rate-matched CEM, colored noise alone solves 6 Push-T episodes against 28 (p < 0.001) and 4 Wall episodes against 25 (p < 0.001), and the two are level on PointMaze, 3 against 2 (p = 1.0).

\subsection{The scale of the step means}

Section 3.4 lists three changes that come with rate-matched CEM. The step means receive all of the exploration, they start at a larger scale, and the search space is smaller. We separate the scale from the other two on DDP-WM by starting the proposal of rate-matched CEM at standard deviation 0.45 instead of 1.0, the step-mean scale of the official proposal in Table 1. The search stays in S, so the share and the dimension are those of rate-matched CEM, and only the scale is the official one. \emph{At the official scale the gain disappears.} This variant fails in 27, 11, and 6 episodes on Push-T, Wall, and PointMaze (Table 5), against 19, 11, and 6 for the official planner and 14, 2, and 2 for rate-matched CEM. On Wall and PointMaze it fails exactly as often as the official planner, and on Push-T it fails more often. A small space whose exploration lies wholly on the step means does not by itself avoid the failures of Table 2. In these runs the gain comes with step means explored at the scale of a single decision, which per-action sampling reaches only with within-step deviations of standard deviation 2 (Section 3.3).

\begin{table}[t]
\centering\footnotesize\setlength{\tabcolsep}{3pt}
\setcounter{table}{4}
\caption{Rate-matched CEM on DDP-WM with its proposal started at the official step-mean scale, success rate (\%) with failed episodes in parentheses, 150 episodes per setting.}
\begin{tabular}{@{}>{\raggedright\arraybackslash}p{0.2400\dimexpr\linewidth-6\tabcolsep\relax}>{\raggedright\arraybackslash}p{0.2533\dimexpr\linewidth-6\tabcolsep\relax}>{\raggedright\arraybackslash}p{0.2533\dimexpr\linewidth-6\tabcolsep\relax}>{\raggedright\arraybackslash}p{0.2533\dimexpr\linewidth-6\tabcolsep\relax}@{}}
\toprule
Environment & official & held, std 1.0 & held, std 0.45 \\
\midrule
Push-T & 87.3 (19) & 90.7 (14) & 82.0 (27) \\
Wall & 92.7 (11) & 98.7 (2) & 92.7 (11) \\
PointMaze & 96.0 (6) & 98.7 (2) & 96.0 (6) \\
\bottomrule\end{tabular}
\end{table}

\begin{figure}[t]\centering

\includegraphics[width=\linewidth]{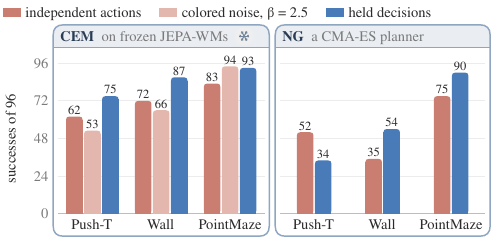}

\setcounter{figure}{4}

\caption{JEPA-WMs, successes out of 96 per benchmark. Left: CEM with the official proposal, with the colored noise of iCEM (\ensuremath{\beta} = 2.5), and with rate-matched decisions. Right: NG with independent and with held decisions.}\end{figure}

\section{Other planners and other world models}

The bound needs only one good member, but a practitioner will ask what held decisions do for other planners and other models. We gave held decisions, one sampled action per model step repeated for the model's frameskip, to a planner that executes a single sample and to four world models that come with planners of their own.

\subsection{A planner that executes a single sample}

CEM hands the controller the mean of its elites. NG, the Nevergrad-based planner that JEPA-WMs introduces~\cite{ref38} and that here runs CMA-ES~\cite{ref19}, hands it the single candidate that the model scores best. On JEPA-WMs, held decisions raise NG's success on Wall from 35 to 54 of 96 episodes (33 solved only when held and 14 only when not, p = 0.008) and on PointMaze from 75 to 90 (20 against 5, p = 0.004; that baseline ran under another version of the cma library, so the pairing is not exact). On Push-T they lower it from 52 to 34 (6 against 24, p = 0.001). Table 6 measures why on the first 24 Push-T tasks.

\begin{table}[t]
\centering\footnotesize\setlength{\tabcolsep}{3pt}
\setcounter{table}{5}
\caption{JEPA-WMs on the first 24 Push-T tasks. Successes, the median cost that the model predicts for the returned plan and the median cost of the state the plan reaches, both under the official cost, and the fraction of five-step net displacements outside the range of the training data.}
\begin{tabular}{@{}>{\raggedright\arraybackslash}p{0.1200\dimexpr\linewidth-12\tabcolsep\relax}>{\raggedright\arraybackslash}p{0.1950\dimexpr\linewidth-12\tabcolsep\relax}>{\raggedright\arraybackslash}p{0.1650\dimexpr\linewidth-12\tabcolsep\relax}>{\raggedright\arraybackslash}p{0.1650\dimexpr\linewidth-12\tabcolsep\relax}>{\raggedright\arraybackslash}p{0.1550\dimexpr\linewidth-12\tabcolsep\relax}>{\raggedright\arraybackslash}p{0.2000\dimexpr\linewidth-12\tabcolsep\relax}@{}}
\toprule
Planner & Actions & Successes & Predicted cost & Reached cost & Outside training range \\
\midrule
CEM & independent & 17 & 0.089 & 0.147 & 0\% \\
CEM & held & 21 & 0.075 & 0.081 & 0.7\% \\
NG & independent & 11 & 0.123 & 0.278 & 5.6\% \\
NG & held & 6 & 0.132 & 0.535 & 35.4\% \\
\bottomrule\end{tabular}
\end{table}

A held sample moves all five primitive actions of a step in one direction, and Push-T does not clip actions, so a single held candidate can carry the plan past the training data in one step, where the model scores it best and the episode fails. For every returned plan we also recorded the cost that the model predicts and the cost of the state that the plan reaches, under the same encoder. For held CEM plans the two agree, with medians of 0.075 and 0.081. For held NG plans the reached cost is four times the predicted one, 0.535 against 0.132, and the distance between the predicted and the reached latent state has a median of 0.545 in the failed episodes of this row against 0.101 in the successful ones. Over the 24 episodes of this row, the mean size of the planned actions and this distance have a rank correlation of 0.77, against 0.25 to 0.44 in the other three rows. The model underrates these plans, and they are the plans that leave the training range. This is the second way of Section 2, now measured. The CEM rows settle two further points. The released planner already stays within the training range, so rate-matched CEM does not win by pulling plans back into it; and an elite mean keeps held plans inside the range where a single held sample does not. Held decisions therefore suit planners that execute a refitted mean, which is what the released planners of these models do.

\subsection{Four further world models}

The temporally straightened model of Wang et al.~\cite{ref40} runs in its official configuration, open loop toward goals 25 steps ahead, under CEM and gradient descent on Wall, UMaze, and Push-T. Scope-WM~\cite{refscope}, with 32 or 98 latent tokens, runs on Push-T under CEM and EB-CEM, its elite-bank CEM, which stores the best plans of the first planning round and draws most later candidates around them (Appendix C.5). JEPA-Bisim~\cite{reftoso} releases a DINO-Bisim model and a DINO-WM trained on 1,000 Push-T trajectories, and C-JEPA~\cite{refnam} predicts over object-centric representations with blocks of five actions. These three run on Push-T under CEM in their official configurations, and their released planners reproduce the reported success rates to within one episode.

\begin{table*}[t]
\centering\footnotesize\setlength{\tabcolsep}{3pt}
\setcounter{table}{6}
\caption{Successes with the official proposal and with held decisions, and the two-sided paired p value; 150 episodes per cell for the straightened model and C-JEPA, 50 for Scope-WM and the JEPA-Bisim models.}
\begin{tabular}{@{}>{\raggedright\arraybackslash}p{0.3000\dimexpr\linewidth-8\tabcolsep\relax}>{\raggedright\arraybackslash}p{0.1750\dimexpr\linewidth-8\tabcolsep\relax}>{\raggedright\arraybackslash}p{0.1750\dimexpr\linewidth-8\tabcolsep\relax}>{\raggedright\arraybackslash}p{0.1750\dimexpr\linewidth-8\tabcolsep\relax}>{\raggedright\arraybackslash}p{0.1750\dimexpr\linewidth-8\tabcolsep\relax}@{}}
\toprule
Model & Planner & Environment & Official \ensuremath{\to} held & p \\
\midrule
straightened & CEM & Wall & 149 \ensuremath{\to} 144 & n.s. \\
straightened & CEM & UMaze & 137 \ensuremath{\to} 147 & 0.006 \\
straightened & CEM & Push-T & 101 \ensuremath{\to} 109 & n.s. \\
straightened & gradient descent & Wall & 141 \ensuremath{\to} 140 & n.s. \\
straightened & gradient descent & UMaze & 125 \ensuremath{\to} 130 & n.s. \\
straightened & gradient descent & Push-T & 106 \ensuremath{\to} 112 & n.s. \\
Scope-WM, 32 tokens & CEM & Push-T & 42 \ensuremath{\to} 44 & 0.69 \\
Scope-WM, 32 tokens & EB-CEM & Push-T & 43 \ensuremath{\to} 45 & 0.63 \\
Scope-WM, 98 tokens & CEM & Push-T & 45 \ensuremath{\to} 48 & 0.45 \\
Scope-WM, 98 tokens & EB-CEM & Push-T & 47 \ensuremath{\to} 48 & 1.0 \\
DINO-Bisim & CEM & Push-T & 18 \ensuremath{\to} 19 & 1.0 \\
DINO-WM, 1,000 trajectories & CEM & Push-T & 23 \ensuremath{\to} 28 & 0.23 \\
C-JEPA & CEM & Push-T & 133 \ensuremath{\to} 133 & 1.0 \\
\bottomrule\end{tabular}
\end{table*}

Held decisions raise success in ten of the thirteen settings, do not change it in one, and lower it significantly in none. The one significant change is on UMaze, where they remove 10 of the 13 failures of CEM; the others are small changes on 150 or 50 episodes. On the 32-token Scope-WM, plain CEM with held decisions solves 44 of 50 episodes, the 88\% that the Scope-WM paper reports for its EB-CEM. The count does not move on C-JEPA, which solves 133 of 150 episodes under either proposal. Gradient descent, which follows the cost gradient rather than sampling, gains on UMaze and Push-T, and the two losses are on Wall, where the official planners already succeed in 149 and 141 of 150 episodes. Together with the nine settings of Section 4, held decisions raise success in 19 of 22 settings under CEM-type and gradient planners and never lower it significantly.

\section{Related work}

Planning with learned world models. Learning a model of the environment and planning in it reaches from recurrent world models~\cite{ref14} and latent dynamics models for control~\cite{ref15,ref16,ref18,ref36} to video and navigation world models that plan with CEM~\cite{ref2,ref3}. The block-input models studied here improve the predictor~\cite{ref42}, compare design choices for JEPA world models~\cite{ref38}, or shape the latent space for planning~\cite{ref40,reftoso,refnam}. In each case the comparison is made under the planner the model adopts, which is the practice our bound reads. Hamrick et al.~\cite{ref17} ask what planning contributes to model-based reinforcement learning, and Lambert et al.~\cite{ref25} find that a model's training objective and its value for control can disagree; we hold the model fixed and bound how much of its reported failure the planner adds.

How sampling planners explore. CEM~\cite{ref34,ref10} and path-integral control~\cite{ref41} are the default optimizers for learned dynamics~\cite{ref8}, and gradient-based planners descend the same costs~\cite{ref4}. Refinements change how a given space is sampled, each with parameters of its own: colored noise with an elite memory~\cite{ref33}, filtered or low-frequency noise~\cite{ref31,ref39}, spline knots~\cite{ref21}, and covariance adaptation~\cite{ref19}. Rate-matched CEM instead changes which space is searched and takes that space from training. We set it beside spline-style interpolation (Section 5.1), colored noise (Section 5.2), and a covariance-adapting planner (Section 6.1).

Holding actions. Frame skipping~\cite{ref29} and action repeat~\cite{ref15} hold an action for several environment steps so that an agent acts less often; learned repetition chooses the duration~\cite{ref24,ref35}; action chunking predicts blocks of future actions~\cite{ref43,ref7}; and move blocking fixes the input over groups of steps to make a receding-horizon problem smaller, at some cost in performance~\cite{ref6}. Each treats the hold length as a design choice for control. In the models studied here training has already set it, and the released planner does not use it. Two of our results separate rate-matched CEM from holding as such: its interpolated variant changes the action at every environment step and keeps the gain (Section 5.1), while holding the single sample of NG costs success on Push-T (Section 6.1).

Search error and re-evaluation. In machine translation, exact and optimal decoding measured how much of a system's error came from search~\cite{ref13,ref37}. Our split follows that work and mirrors the decomposition of excess error in learning, with the planner in the place of the optimizer~\cite{ref5}; planners that exploit the errors of learned models~\cite{ref23,ref22} form its second way, which Section 6.1 measures. Controlled re-evaluations have changed how several fields read their numbers~\cite{ref28,ref26,ref20,ref30,ref1}. This one comes with a planner that improves the numbers it re-evaluates.

\section{Conclusion}

A reported success rate belongs to a model and its planner. For DINO-WM, DDP-WM, and JEPA-WMs, a planner that uses nothing beyond the model, its cost, and the budget avoids 74 of 135 reported failures, a lower bound of 55\% on the share of search error. That planner, rate-matched CEM, follows from how the models are trained, has no parameter, needs no retraining, and gives reference results above the best official ones. After two planning rounds it fails less often than the released planner after three. Its gain survives interpolation and late refinement, is not reproduced by colored noise, disappears when its decisions start at the step-mean scale of the released proposal, and appears in 19 of 22 settings with CEM-type and gradient planners; the one significant loss comes from a planner whose single held samples leave the training range.

The bounds refer to the official costs, budgets, and benchmarks, which are the numbers this line of work reports. Running rate-matched CEM beside the released planner costs one further evaluation and separates what a model can do from what its planner finds.

\FloatBarrier\clearpage
\includepdf[pages=-]{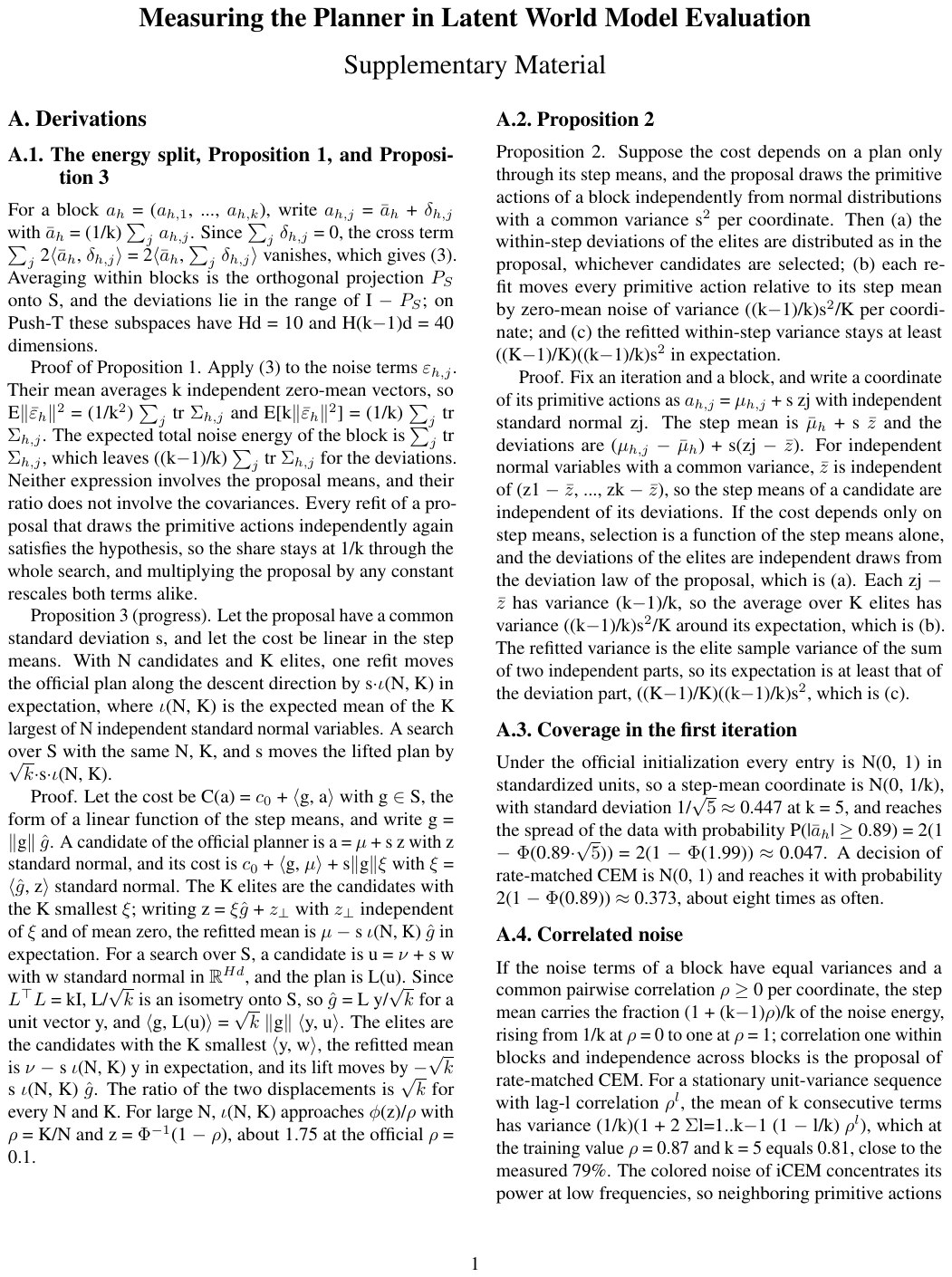}

{\small\begin{thebibliography}{99}
\bibitem{ref1} Rishabh Agarwal, Max Schwarzer, Pablo Samuel Castro, Aaron Courville, and Marc G. Bellemare. Deep reinforcement learning at the edge of the statistical precipice. In NeurIPS, 2021.

\bibitem{ref2} Mido Assran, Adrien Bardes, David Fan, Quentin Garrido, et al. V-JEPA 2: Self-supervised video models enable understanding, prediction and planning. arXiv preprint arXiv:2506.09985, 2025.

\bibitem{ref3} Amir Bar, Gaoyue Zhou, Danny Tran, Trevor Darrell, and Yann LeCun. Navigation world models. In CVPR, 2025.

\bibitem{ref4} Homanga Bharadhwaj, Kevin Xie, and Florian Shkurti. Model-predictive control via cross-entropy and gradient-based optimization. In Learning for Dynamics and Control, 2020.

\bibitem{ref5} Léon Bottou and Olivier Bousquet. The tradeoffs of large scale learning. In NeurIPS, 2007.

\bibitem{ref6} Raphael Cagienard, Pascal Grieder, Eric C. Kerrigan, and Manfred Morari. Move blocking strategies in receding horizon control. Journal of Process Control, 17(6):563--570, 2007.

\bibitem{ref7} Cheng Chi, Siyuan Feng, Yilun Du, Zhenjia Xu, Eric Cousineau, Benjamin Burchfiel, and Shuran Song. Diffusion policy: Visuomotor policy learning via action diffusion. In RSS, 2023.

\bibitem{ref8} Kurtland Chua, Roberto Calandra, Rowan McAllister, and Sergey Levine. Deep reinforcement learning in a handful of trials using probabilistic dynamics models. In NeurIPS, 2018.

\bibitem{ref9} William G. Cochran. Some methods for strengthening the common \ensuremath{\chi}\ensuremath{{}^2} tests. Biometrics, 10(4):417--451, 1954.

\bibitem{ref10} Pieter-Tjerk de Boer, Dirk P. Kroese, Shie Mannor, and Reuven Y. Rubinstein. A tutorial on the cross-entropy method. Annals of Operations Research, 134(1):19--67, 2005.

\bibitem{ref11} Pete Florence, Corey Lynch, Andy Zeng, Oscar A. Ramirez, Ayzaan Wahid, Laura Downs, Adrian Wong, Johnny Lee, Igor Mordatch, and Jonathan Tompson. Implicit behavioral cloning. In CoRL, PMLR 164:158--168, 2021.

\bibitem{ref12} Justin Fu, Aviral Kumar, Ofir Nachum, George Tucker, and Sergey Levine. D4RL: Datasets for deep data-driven reinforcement learning. arXiv preprint arXiv:2004.07219, 2020.

\bibitem{ref13} Ulrich Germann, Michael Jahr, Kevin Knight, Daniel Marcu, and Kenji Yamada. Fast decoding and optimal decoding for machine translation. In ACL, pages 228--235, 2001.

\bibitem{ref14} David Ha and Jürgen Schmidhuber. Recurrent world models facilitate policy evolution. In NeurIPS, 2018.

\bibitem{ref15} Danijar Hafner, Timothy Lillicrap, Ian Fischer, Ruben Villegas, David Ha, Honglak Lee, and James Davidson. Learning latent dynamics for planning from pixels. In ICML, 2019.

\bibitem{ref16} Danijar Hafner, Timothy Lillicrap, Jimmy Ba, and Mohammad Norouzi. Dream to control: Learning behaviors by latent imagination. In ICLR, 2020.

\bibitem{ref17} Jessica B. Hamrick, Abram L. Friesen, Feryal Behbahani, Arthur Guez, Fabio Viola, Sims Witherspoon, Thomas Anthony, Lars Buesing, Petar Veličković, and Théophane Weber. On the role of planning in model-based deep reinforcement learning. In ICLR, 2021.

\bibitem{ref18} Nicklas Hansen, Xiaolong Wang, and Hao Su. Temporal difference learning for model predictive control. In ICML, 2022.

\bibitem{ref19} Nikolaus Hansen and Andreas Ostermeier. Completely derandomized self-adaptation in evolution strategies. Evolutionary Computation, 9(2):159--195, 2001.

\bibitem{ref20} Peter Henderson, Riashat Islam, Philip Bachman, Joelle Pineau, Doina Precup, and David Meger. Deep reinforcement learning that matters. In AAAI, 2018.

\bibitem{ref21} Taylor Howell, Nimrod Gileadi, Saran Tunyasuvunakool, Kevin Zakka, Tom Erez, and Yuval Tassa. Predictive sampling: Real-time behaviour synthesis with MuJoCo. arXiv preprint arXiv:2212.00541, 2022.

\bibitem{ref22} Michael Janner, Justin Fu, Marvin Zhang, and Sergey Levine. When to trust your model: Model-based policy optimization. In NeurIPS, 2019.

\bibitem{ref23} Thanard Kurutach, Ignasi Clavera, Yan Duan, Aviv Tamar, and Pieter Abbeel. Model-ensemble trust-region policy optimization. In ICLR, 2018.

\bibitem{ref24} Aravind S. Lakshminarayanan, Sahil Sharma, and Balaraman Ravindran. Dynamic action repetition for deep reinforcement learning. In AAAI, 2017.

\bibitem{ref25} Nathan Lambert, Brandon Amos, Omry Yadan, and Roberto Calandra. Objective mismatch in model-based reinforcement learning. In Learning for Dynamics and Control, PMLR 120:761--770, 2020.

\bibitem{refscope} Chunzheng Li, Zesheng Jia, Hongda Zhang, Jiaying Tang, Yuntian Wang, Siao Liu, and Jin Wang. Scope-WM: Scoped computation for efficient visual world models. arXiv preprint arXiv:2609.33218, 2026.

\bibitem{ref26} Mario Lucic, Karol Kurach, Marcin Michalski, Sylvain Gelly, and Olivier Bousquet. Are GANs created equal? A large-scale study. In NeurIPS, 2018.

\bibitem{ref27} Nathan Mantel and William Haenszel. Statistical aspects of the analysis of data from retrospective studies of disease. Journal of the National Cancer Institute, 22(4):719--748, 1959.

\bibitem{ref28} Gábor Melis, Chris Dyer, and Phil Blunsom. On the state of the art of evaluation in neural language models. In ICLR, 2018.

\bibitem{ref29} Volodymyr Mnih, Koray Kavukcuoglu, David Silver, et al. Human-level control through deep reinforcement learning. Nature, 518(7540):529--533, 2015.

\bibitem{ref30} Kevin Musgrave, Serge Belongie, and Ser-Nam Lim. A metric learning reality check. In ECCV, 2020.

\bibitem{ref31} Anusha Nagabandi, Kurt Konolige, Sergey Levine, and Vikash Kumar. Deep dynamics models for learning dexterous manipulation. In CoRL, PMLR 100:1101--1112, 2019.

\bibitem{refnam} Heejeong Nam, Quentin Le Lidec, Lucas Maes, Yann LeCun, and Randall Balestriero. Causal-JEPA: Learning world models through object-level latent masking. In ICML, 2026.

\bibitem{ref32} Maxime Oquab, Timothée Darcet, Théo Moutakanni, et al. DINOv2: Learning robust visual features without supervision. Transactions on Machine Learning Research, 2024.

\bibitem{ref33} Cristina Pinneri, Shambhuraj Sawant, Sebastian Blaes, Jan Achterhold, Joerg Stueckler, Michal Rolinek, and Georg Martius. Sample-efficient cross-entropy method for real-time planning. In CoRL 2020, PMLR 155:1049--1065, 2021.

\bibitem{ref34} Reuven Y. Rubinstein. The cross-entropy method for combinatorial and continuous optimization. Methodology and Computing in Applied Probability, 1(2):127--190, 1999.

\bibitem{ref35} Sahil Sharma, Aravind S. Lakshminarayanan, and Balaraman Ravindran. Learning to repeat: Fine grained action repetition for deep reinforcement learning. In ICLR, 2017.

\bibitem{ref36} Uladzislau Sobal, Wancong Zhang, Kyunghyun Cho, Randall Balestriero, Tim G. J. Rudner, and Yann LeCun. Learning from reward-free offline data: A case for planning with latent dynamics models. In NeurIPS, 2025.

\bibitem{ref37} Felix Stahlberg and Bill Byrne. On NMT search errors and model errors: Cat got your tongue? In EMNLP-IJCNLP, 2019.

\bibitem{ref38} Basile Terver, Tsung-Yen Yang, Jean Ponce, Adrien Bardes, and Yann LeCun. What drives success in physical planning with joint-embedding predictive world models? Transactions on Machine Learning Research, 2026.

\bibitem{reftoso} Leonardo F. Toso, Davit Shadunts, Yunyang Lu, Gloria Geng, Nihal Sharma, Donglin Zhan, Nam H. Nguyen, and James Anderson. JEPA-Bisim: Learning robust visual representations for planning with joint-embedding predictive world models. arXiv preprint arXiv:2602.18639, 2026.

\bibitem{ref39} Bogdan Vlahov, Jason Gibson, David D. Fan, Patrick Spieler, Ali-akbar Agha-mohammadi, and Evangelos A. Theodorou. Low frequency sampling in model predictive path integral control. IEEE Robotics and Automation Letters, 9(5):4543--4550, 2024.

\bibitem{ref40} Ying Wang, Oumayma Bounou, Gaoyue Zhou, Randall Balestriero, Tim G. J. Rudner, Yann LeCun, and Mengye Ren. Temporal straightening for latent planning. In ICML, 2026.

\bibitem{ref41} Grady Williams, Andrew Aldrich, and Evangelos A. Theodorou. Model predictive path integral control: From theory to parallel computation. Journal of Guidance, Control, and Dynamics, 40(2):344--357, 2017.

\bibitem{ref42} Shicheng Yin, Kaixuan Yin, Weixing Chen, Yang Liu, Guanbin Li, and Liang Lin. DDP-WM: Disentangled dynamics prediction for efficient world models. In ICML, 2026.

\bibitem{ref43} Tony Z. Zhao, Vikash Kumar, Sergey Levine, and Chelsea Finn. Learning fine-grained bimanual manipulation with low-cost hardware. In RSS, 2023.

\bibitem{ref44} Gaoyue Zhou, Hengkai Pan, Yann LeCun, and Lerrel Pinto. DINO-WM: World models on pre-trained visual features enable zero-shot planning. In ICML, 2025.
\end{thebibliography}}
\end{document}